\documentclass[journal]{./support/IEEEtran}

\IEEEoverridecommandlockouts
\usepackage{cite}
\usepackage{amsmath,amssymb,amsfonts,mathtools}
\usepackage[linesnumbered,ruled,vlined]{algorithm2e}
\usepackage{graphicx}
\usepackage{textcomp}
\usepackage{xcolor}
\usepackage{tensor}
\usepackage{float}
\usepackage{nomencl}
\makenomenclature
\usepackage{subcaption}
\usepackage[OT1]{fontenc} 
\usepackage{multirow}
\usepackage{gensymb}
\usepackage{hyperref}

\newlength{\imagem}

\newlength{\images}

\graphicspath{{./figs/}}
\DeclareGraphicsExtensions{.png,.jpg,.eps,.pdf}

\title{BDPGO: Balanced Distributed Pose Graph Optimization Framework for Swarm Robotics}

\author{Hao Xu, Shaojie Shen
\thanks{All authors are with the Department of Electronic and Computer Engineering, Hong Kong University of Science and Technology, Hong Kong, China.
{\tt\small hxubc@connect.ust.hk, eeshaojie@ust.hk} 
}
}

\begin{document}
\maketitle
\begin{abstract}
Distributed pose graph optimization (DPGO) is one of the fundamental techniques of swarm robotics. 
Currently, the sub-problems of DPGO are built on the native poses.
Our validation proves that this approach may introduce an imbalance in the sizes of the sub-problems in real-world scenarios, which affects the speed of DPGO optimization, and potentially increases communication requirements. 
In addition, the coherence of the estimated poses is not guaranteed when the robots in the swarm fail, or partial robots are disconnected.
In this paper, we propose BDPGO, a balanced distributed pose graph optimization framework using the idea of decoupling the robot poses and DPGO.
BDPGO re-distributes the poses in the pose graph to the robot swarm in a balanced way by introducing a two-stage graph partitioning method to build balanced subproblems.
Our validation demonstrates that BDPGO significantly improves the optimization speed without changing the specific algorithm of DPGO in realistic datasets. 
What's more, we also validate that BDPGO is robust to robot failure, changes in the wireless network. BDPGO has capable of keeps the coherence of the estimated poses in these situations.
The framework also has the potential to be applied to other collaborative simultaneous localization and mapping (CSLAM) problems involved in distributedly solving the factor graph.
\end{abstract}

\section{Introduction}\label{sect:intro}

One of the fundamental problems for autonomous swarm robotics is collaborative simultaneous localization and mapping (CSLAM)\cite{michael2014collaborative, lajoie2020door, zou2019collaborative}.
In CSLAM problems, consistency across the swarm and real-time performance are often more important than local accuracy. 
Hence, CSLAM systems often use methods such as visual-inertial odometry (VIO)\cite{qin2018vins} and lidar-inertial odometry (LIO)\cite{zhang2014} to estimate the ego-motion while using pose graph optimization (PGO) to estimate the state of the entire swarm to ensure consistency.
At an earlier time, researchers tried to solve PGO in a centralized manner \cite{michael2014collaborative}.
However, the computation power required for centralized PGO snowballs as the swarm's scale grows, leading to challenges in building a large-scale robot swarm.

Recently, distributed pose graph optimization (DPGO) has become an increasingly popular research direction\cite{tian2021distributed, tian2020asynchronous,choudhary2017distributed}.
These methods adopt distributed optimization\cite{bertsekas1989parallel, tian2020asynchronous, tian2021distributed} to solve the PGO and have made considerable progress on addressing the scalability issue.
The general idea of distributed optimization \cite{bertsekas1989parallel,tsianos2012push} is to partition the global optimization problem into multiple sub-problems, set up the links between sub-problems.
During the optimization procedure, the robots need to solve the sub-problems individually as well as exchange data between themselves to obtain the solution of the global optimization problem iteratively.

In DPGO systems, keyframes are extracted by the VIO or LIO, while loop detection methods\cite{lajoie2020door} are adopted to detect loops, namely relative poses between keyframes. 
The poses attached to keyframes are represented as the vertices in the pose graph \cite{qin2018vins,mur2015orb} and the detected loops are represented as edges.
A straightforward idea in current works is to build the sub-problems on each robot with the poses attached to native keyframes.
This straightforward idea is treated as the baseline method in this paper\cite{choudhary2017distributed,chang2020kimera}.

For swarm robots in real-world applications, however, the movements of robots are imbalanced when accomplishing various types of tasks. 
Therefore, the sizes of the sub-problem generated by each robot with the baseline approach can be highly imbalanced among the swarm,
for example, when a single robot in the swarm performs a specific task, and the remaining robots are idle. 
Intuitively, this imbalance in the sizes of the sub-problems will produce an imbalance in the speed of solving each sub-problem and thus affects the utilization of computation power.
In particular, for synchronous DPGO algorithms \cite{choudhary2017distributed}, the biggest sub-problem determines the global optimization speed,
while for asynchronous DPGO algorithms\cite{tian2020asynchronous}, the speed of solving the imbalanced sub-problems still affects the convergence properties, as will be verified in Sect.\ref{sect:evaluation}.

Furthermore, the volume of data to be exchanged when solving the DPGO is determined by the number of keyframes involved in the inter-edges.
Our evaluation results in Sect. \ref{sect:expr_dpgo} show that the baseline approach may occupy more communication volume in exchanging poses among the swarm compared to the optimal sub-problem partitioning.
What's more, the wireless networks of robot swarms are complex and often unstable.
Some robots use wireless ad hoc networks to maintain communication\cite{chung2018survey} and may experience partial robots are disconnected.
However, with the baseline method, the robot swarm will lose the poses and edges of the robots that disconnected or failed, so the coherence of the estimated poses is not guaranteed.

Looking to other areas where distributed computing has been applied, for example, computational fluid dynamics (CFD), space is divided into huge meshes, which can be considered as graph structures.
In the process of solving partial differential equations (PDEs) on these meshes, researchers  distribute the tasks to processers by cutting the graph with graph partitioning techniques\cite{hendrickson2000graph} to solve the problem in parallel.
A similar approach is applicable to DPGO, in which we can balance and redistribute the poses and the sub-problems by partitioning the pose graph.
Proper partitioning of the pose graph can thus deliver balanced sub-problems with lower communication volume, which can significantly improve the performance of DPGO, including speeding up the optimization, accelerating the convergence of the global problem, and reducing communication bandwidth, as we will verify in Sect. \ref{sect:evaluation}.

However, in a practical CSLAM system, the pose graph grows with time, yet real-time performance is important for such systems.
This makes it inappropriate to directly apply typical graph partitioning methods\cite{karypis1997metis}, which take a long time for each partition, affecting the real-time performance of the system and also breaking the decentralized and distributed nature of CSLAM systems.
Therefore, in this paper, we propose a \textit{balanced distributed pose graph optimization (BDPGO)} framework to maintain balanced sub-problem sizes in distributed optimization while minimizing the communication volume between robots.
The basic idea of this approach is to decouple the measurements and computing functions of the robots by redistributing the poses and edges with a graph partitioning method.

BDPGO involves a two-stage graph partitioning approach that combines the streaming graph partitioning and repartitioning techniques to balance the overhead and quality of graph partitioning for the pose graph.
In addition, our framework is designed to be decentralized while being capable of responding to various changes in the swarm, including changes in the wireless network and the failure of robots.

To summarize, the main contributions of this paper are as follows:
\begin{itemize}
    \item We propose a general framework of balanced distributed pose graph optimization with two-stage incremental graph partitioning (BDPGO).
    The proposed framework applies to various existing DPGO algorithms with a small overhead but significant improvements. 
    Furthermore, the proposed framework is also robust to robot failures in the swarm or network topology changes.
    \item We evaluate the impact of graph partitioning on the DPGO with two categories of DPGO approaches. 
    The results show that proper pose graph partitioning with the proposed method speeds up the optimization, accelerating convergence, and reduces the communication requirements significantly compared to the baseline.
\end{itemize}

\section{Related Works}\label{sect:related}
\subsection{Distributed Pose Graph Optimization}
PGO\cite{Rosen2019SESync,briales2017cartan}, derived from the factor graph\cite{dellaert2017factor}, is one of the key techiques for simultaneous localization and mapping (SLAM), and is generally used for re-localization\cite{qin2018vins, mur2015orb} and dense mapping\cite{reijgwart2020voxgraph}.
When researchers turned their attention to multi-robot collaboration, the PGO approach was naturally introduced in CSLAM as well\cite{cunningham2010ddf, cunningham2013ddf,  michael2014collaborative}.
Early attempts at PGO in CSLAM were accomplished by solving the pose graph problem using a centralized server\cite{michael2014collaborative}. However, centralized PGO suffers from scalability and communication issues in swarm robot systems.
The first distributed SLAM approach is proposed by Cunningham et al.\cite{cunningham2010ddf}, which uses a constrained factor graph in DDF-SAM, as well as its improved version DDF-SAM2\cite{cunningham2013ddf}.
However, both methods keep a neighborhood graph on each agent and optimize it on each agent, which wastes computational resources and lacks scalability.

Real progress on DPGO started with the Distributed Gauss-Seidel (DGS) approach proposed by Choudhary et al.\cite{choudhary2017distributed}.
In this method, the PGO problem is first transformed into a linear problem and then solved in a distributed manner using the Gauss-Seidel method\cite{bertsekas1989parallel}.
Later the ASAPP method, based on distributed gradient descent, was proposed by Tian et al.\cite{tian2020asynchronous}. It can be seen as a distributed and asynchronous version of Riemann gradient descent\cite{boumal2020introduction}. The highlight of that work is that it was the first to take communication into account and deliver an asynchronous DPGO approach. 
The subsequent work of Tian et al.\cite{tian2021distributed} introduces a DC2-PGO method using the Distributed Riemann-Staircase method that is a certifiably correct DPGO method.
The above  DPGO methods are currently used in several practical SLAM systems, including DOOR-SLAM\cite{lajoie2020door}, which utilizes DGS, and Kimera-multi\cite{chang2020kimera}, which uses ASAPP.
However, none of these existing systems takes into account the problem of sub-problem partitioning. 
In addition, robot failure and changes in the wireless network have not been considered.

\subsection{Graph Parititioning}
Graph partitioning techniques have a decades-long history, and one of the major reasons that such methods have been developed is to optimize computational performance and communication in distributed computing \cite{schloegel2000unified,hendrickson2000graph, schloegel1997multilevel}.
Graph partitioning methods are used to deal with meshing problems in numerical computations to improve the parallelized processing of numerical computations\cite{schloegel1997multilevel, schloegel2000unified}.

Two main approaches exist to split the graph, edge-cut partitioning\cite{lasalle2013multi} and vertex-cut partitioning\cite{gonzalez2012powergraph}, and these two branches of partitioning 
algorithms have been very thoroughly studied.
However,  early algorithms, such as METIS\cite{lasalle2013multi}, were designed offline. As technology has evolved, the size of data is now often enormous, so researchers have begun to consider streaming graph partitioning (SGP) for both edge-cut and vertex-cut parititioniong\cite{stanton2012streaming, tsourakakis2014fennel,gonzalez2012powergraph}, in which the graph partitioning is performed in real-time as the graph structures are received as an input stream. 
In general, these streaming approaches can be computed in a distributed manner.

Besides the stream-based approach, researchers are taking another type of problem into consideration, graph repartitioning\cite{schloegel2000unified}.
The graph repartitioning method is designed to improve an existing partition, taking into account not only the edge-cut of the partitions and balance between partitionings but also the cost of data redistribution.
\subsection{Graph Parititioning for Distributed Optimization}

The utilization of graph partitioning methods to improve distributed optimization is not common.
One example, though, is Guo et al.\cite{guo2015intelligent}, who proposed a spectral clustering-based method for partitioning the sub-problems of distributed optimization, which is used to solve the ac optimal power flow problem in the smart grid. 
However, the structure of the smart grid is static, which is fundamentally different from our DPGO problem, where our graph is incrementally growing over time.

The only current work that involves graph partitioning with PGO is by Tang et al.\cite{tang2016pose}.
However, this work does not involve DPGO, and the proposed partitioning method is designed to transform PGO into a multi-layer problem. The subgraphs of the pose graph in this paper are solved independently and based on these subgraphs, and a simplified upper-level graph is generated for optimizing the global pose graph.
\section{Preliminaries}
\subsection{Graph Partitioning}\label{sect:part}
\begin{figure}[t]
    \centering
    \includegraphics[width=0.4\textwidth]{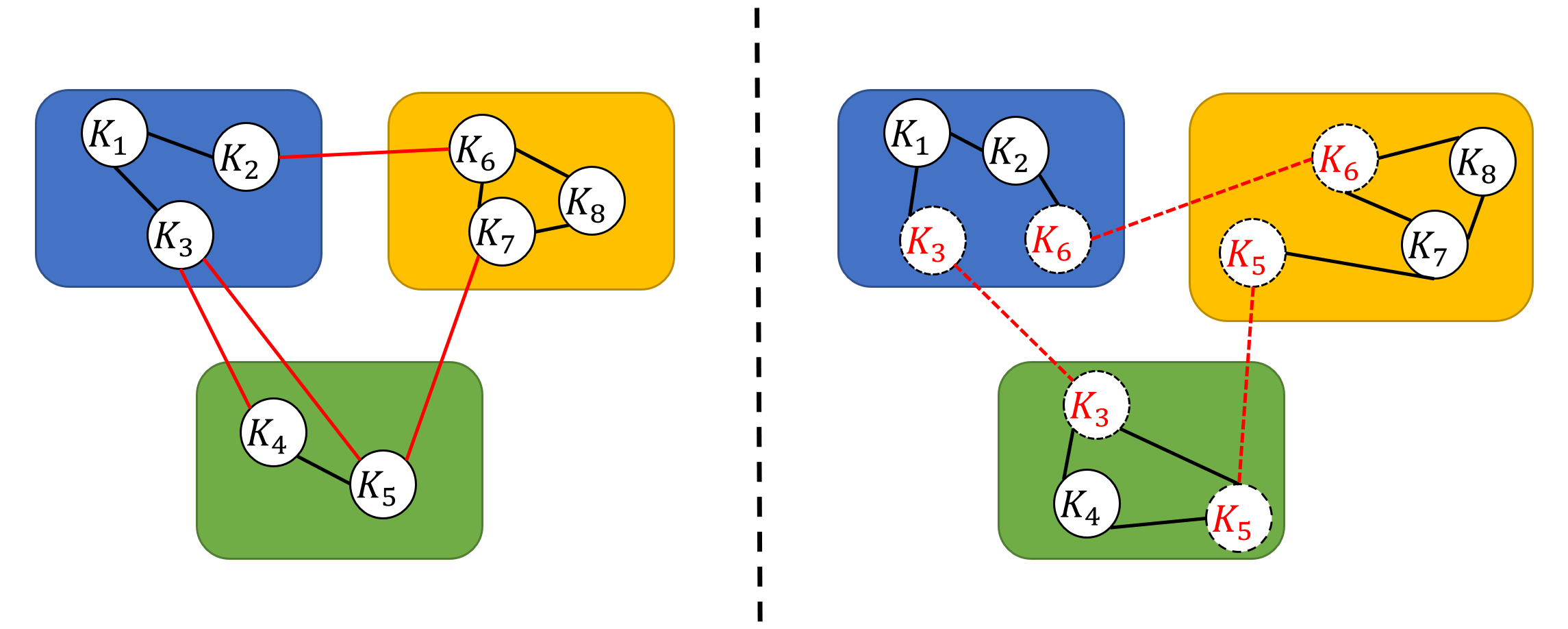}
    \caption{\small{
        This figure demonstrates two approaches to graph partitioning.
        a) Edge-cut partitioning: The vertices graph is partitioned into disjoint sets. 
        The circled $K_1\ ...\ K_8$ denote the vertices.
        The black lines in the graph denote intra-edges inside the partitions, and the red lines represent the inter-edges that cross the partitions. The set of the inter-edges is the edge-cut.
        b) Vertex-cut partitioning. 
        The red circled vertices are the vertex replicas.
    }}\label{fig:edgevsvert}
\end{figure}
Let $G = (V, E)$ denote an  undirected graph, where $V$ is the set of vertices and $E$ is the set of edges. $|V|$ is the number of vertices and $|E|$ is the number of edges.  
A pose graph $G_p$ is a graph in which vertices are the poses  $\mathbf{R}_1^{1}\ \mathbf{p}_1^1\ ...\  \mathbf{R}_n^{l_n} \ \mathbf{p}_n^{l_n} $ attached to the keyframes and edges are the relative poses
 $\tilde{\mathbf{R}}_{i_{t_1}}^{j_{t_1}}, \tilde{\mathbf{t}}_{i_{t_1}}^{j,t_1 }\ ...\ \tilde{\mathbf{R}}_{l_{t_2}}^{m_{t_3}}, \tilde{\mathbf{t}}_{l_{t_2}}^{j,t_3 }  $ between keyframes.
The partitioning of graph $G$ is to split $G$ into $k$ disjoint sets $\mathcal{P}_1\ ... \ \mathcal{P}_n$, and $\mathcal{P} = (\mathcal{P}_1\ ... \ \mathcal{P}_n)$ is the graph partitioning.

In edge-cut partitioning, the vertices of graph $G$ are split into disjoint sets $\mathcal{P}_1\ ...\ \mathcal{P}_n$, as shown in  Fig. \ref{fig:edgevsvert} (a). 
The set of edges which connect vertices in different partitions is called the edge-cut.
Typical edge-cut graph partitioning algorithms are designed to minimize the edge-cut.
Meanwhile, vertex-cut partitioning splits edges into disjoint sets \cite{gonzalez2012powergraph}, so some of the vertices may have replicas in different partitions, as shown in Fig. \ref{fig:edgevsvert} (b).

In this paper, we combine the greedy-based SGP with the graph repartition method in order to balance efficiency and partitioning quality, which will be described in detail in Sect. \ref{sect:framework}.

\subsection{Distributed Pose Graph Optimization}\label{sect:optimization}
Assume there are $n$ robots, the distributed optimization problem for pose graph can then be written as\cite{choudhary2017distributed,tian2020asynchronous}

\begin{equation}\label{eq:DPGO}
    \begin{aligned}
        &\min_{\mathbf{x}_i} \sum_{i=1}^n f_{pgo_i}(\mathbf{x}_i)&& \\
        f_{pgo_i}(\mathbf{x}_i)& = \sum_{ (i_{t_0}, j_{t_1}) \in \mathcal{E}^i} \omega_t^2 \left\Vert \mathbf{p}_{j}^{t_1} - \mathbf{p}_{i}^{t_0} - \mathbf{R}_i^{t_0} \tilde{\mathbf{t}}_{i_{t_0}}^{j,t_1 } \right\Vert + &&\\
        &\frac{\omega_R^2}{2} \left\Vert \mathbf{R}_{j}^{t_1} - \mathbf{R}_{i}^{t_0} \tilde{\mathbf{R}}_{i_{t_0}}^{j_{t_1} } \right\Vert, &&\\
        &\mathbf{x}_i = \left[\mathbf{R}_i^{1}\ \mathbf{p}_i^1\ ..\ \mathbf{R}_i^{l_i} \ \mathbf{p}_i^{l_i} \right]^T &&\\
          &s.t.\ \mathbf{R}_i^{t} = \mathbf{E}^\mathbf{R}_{i_t} \mathbf{z}, \mathbf{p}_i^t = \mathbf{E}^p_{i_t} \mathbf{z}, 
        \mathbf{R}_i^{t}\in SO(3), \mathbf{t}\in \mathbb{R}^3,&&
    \end{aligned}
\end{equation}
where $l_i$ is the number of poses of robot $i$, $\mathbf{R}_i^{t}, \mathbf{p}_i^t$ is the pose of robot $i$ of timestamp $t$, $\mathbf{E}^\mathbf{R}_{i_t},  \mathbf{E}^p_{i_t}$ projects $\mathbf{z}$ to pose  $\mathbf{R}_i^{t}, \mathbf{p}_i^t$,  $\mathbf{x}_i$ is the state vector of robot $i$, $\mathcal{E}^i$ is the set of edges involved in state vector $\mathbf{x}_i$,  and $f_{pgo_i}$ is the sub-problem of robot $i$.

Depending on the iterative behavior, there are currently two kinds of DPGO: synchronous \cite{choudhary2017distributed} and asynchronous\cite{tian2020asynchronous}. 
The synchronous methods need to wait for solutions for other robots before proceeding to the next iteration, while the asynchronous methods have no such restriction.
In this paper, we test a synchronous method, DGS\cite{choudhary2017distributed}, and an asynchronous method, ASAPP\cite{tian2020asynchronous}, as the proposed framework's solver. 
DGS solves the local sub-problem by first converting it into a linear problem and then solving it with Gaussian elimination.
In contrast, ASAPP solves the local sub-problem using Riemann gradient descent.
The reader is referred to \cite{choudhary2017distributed} and \cite{tian2020asynchronous} for the specific implementations of both, which are not elaborated here.

In previous DPGO systems\cite{choudhary2017distributed, chang2020kimera, lajoie2020door}, the state vector $\mathbf{x}_i$ is composed by only local poses. 
In our proposed framework, the poses composing $\mathbf{x}_i$ may extracted from remote robots.
The poses of all robots in the swarm $\mathbf{R}_1^{1},\ \mathbf{p}_1^1,\ ...\ \mathbf{R}_n^{l_n},\ \mathbf{p}_n^{l_n} $ are partitioned to disjoint sets $\mathcal{P}_1\ ...\ \mathcal{P}_n$, which is an edge-cut graph partitioning of pose graph $G_p$.
The new state vector $\mathbf{x}_i^*$ is composed of $\mathcal{P}_i=\{\mathbf{R}_j^{t_1}\ \mathbf{p}_j^{t_1}\ \mathbf{R}_k^{t_2}\ \mathbf{p}_k^{t_2}\  ...\ \mathbf{R}_l^{t_3} \ \mathbf{p}_l^{t_3}\}$,
so Problem. \ref{eq:DPGO} can be rewritten by replacing $\mathbf{x}_i$ with $\mathbf{x}_i^*$, which is defined as
\begin{equation}\label{eq:DPGO2}
    \mathbf{x}_i^* = \left[\mathbf{R}_j^{t_1}\ \mathbf{p}_j^{t_1}\ \mathbf{R}_k^{t_2}\ \mathbf{p}_k^{t_2}\  ...\ \mathbf{R}_l^{t_3} \ \mathbf{p}_l^{t_3} \right]^T.
\end{equation}


\subsection{Communication Implementation}\label{sect:comm}

The communication implementation of the robot swarm is critical to the effectiveness of practical DPGO.
Among current CSLAM systems with DPGO, DOOR-SLAM\cite{lajoie2020door} utilizes the peer-to-peer communication model, which was sufficient for its two-UAV real-world experiments.
ASAPP\cite{tian2020asynchronous}, on the other hand, explores the impact of communication delays on convergence performance.

Robot swarms may involve multiple communication models, but the simplest is communication with infrastructure, e.g., wireless routers, cellular networks, etc.
The communication volume\cite{bulucc2016recent, hendrickson2000graph} for each iteration of distributed optimization can be quantified as
\begin{equation}\label{eq:volume}
    \begin{aligned}
    comm(\mathcal{P}) = \sum_{i=1 .. k} comm_i(\mathcal{P}_i) \\
    comm_i(\mathcal{P}_i) = \sigma_p \sum_{v\in \mathcal{P}_i} |D(v)|,
    \end{aligned}
\end{equation}
where $D(v)$ is the set of the partitions which have vertices connected to vertex $v$ excluding partition $P_i$, $\sigma_p$ is the amount of essential data per pose to exchange in each iteration,
$comm(\mathcal{P})$ is the total communication volume of graph partitioning $\mathcal{P}$.

In some scenarios, swarm robot systems communicate within a wireless ad hoc network\cite{toh2001ad} to remove the need for ground infrastructure. 
One of the problems that can be encountered when deploying DPGO on wireless ad hoc networks is that the topology of the wireless mesh network may change as the robots move, and the communication range of the onboard wireless devices is limited. 
For example, the robot swarm may be split into multiple sub-swarms that cannot communicate with each other.
This requires our graph partitioning algorithm to be flexible in responding to changes in network topology and to be capable of coping with the loss of connectivity of the robot.

\section{Proposed Framework}\label{sect:framework}
In this section, we propose a generic framework of balanced distributed pose graph optimization.
The core of the framework is two-stage incremental graph partitioning:
In the first stage of graph partitioning, we partition the incoming keyframes with  FENNEL\cite{tsourakakis2014fennel}, which is a greedy-based SGP approach that has the  capacity to distributedly run on each robots with a very low overhead.
In the second stage, we adopt the Unified Repartitioning Algorithm (URA)\cite{schloegel2000unified} to update the graph for refining the graph partitioning quality in a decentralized manner and with a lower frequency and dealing with the changes of the wireless network or sudden failure of robots.

\begin{algorithm}[ht]
    \caption{Balanced Distributed Pose Graph Optimization Framework with Two-Stage Incremental Graph Partitioning on Each Robot}\label{alg:framework}
    \SetAlgoLined
    \KwData{Partitions $\mathcal{P} = (\mathcal{P}_1 ...\ \mathcal{P}_n)$, Pose Graph $G_{p_k}$, current robot $k$}
    \KwIn{$I_{cur}$}
    \SetKwProg{Fn}{Function}{}{end function}
    \SetKwFunction{FNewkeyframe}{\textbf{HandleNewKeyframe}}
    \SetKwFunction{FRemotekeyframe}{\textbf{HandleRemoteKeyframe}}
    \SetKwFunction{FRemotePart}{\textbf{HandleRemotePartitions}}
    \SetKwFunction{FLoop}{\textbf{PartitioningThread}}
    \SetKwFunction{FLoop2}{\textbf{DistributedSolveThread}}
    \SetKwFunction{FRepart}{\textbf{Repartition}}
    \SetKwFunction{FSGM}{\textbf{FENNEL}}
    \SetKwFunction{FBroadcast}{\textbf{BroadcastPartition}}
    \SetKwFunction{FNetwork}{\textbf{NetworkChanges}}
    \SetKwFunction{FConnectedRobots}{\textbf{ConnectedRobots}}
    \SetKwFunction{FSyncNewPartition}{\textbf{SyncNewPartition}}
    \SetKwFunction{FSyncFullPartition}{\textbf{PullDiffPosesAndPartitions}}
    \SetKwFunction{FSyncFushPartition}{\textbf{PushPartitions}}
    \SetKwFunction{FFushPartition}{\textbf{PushKeyframeEdges}}
    \SetKwFunction{FNetworkChanges}{\textbf{NetworkChanges}}
    \SetKwFunction{FNew}{\textbf{RobotReconnected}}
    \SetKwFunction{FISMaster}{\textbf{IsMain}}
    \SetKwFunction{FNeedRepart}{\textbf{NeedRepartition}}
    \SetKwFunction{FRepart}{\textbf{UnifiedRepartitioning}}
    \SetKwFunction{FUpdateFENNEL}{\textbf{UpdateFENNELParam}}
    \SetKwFunction{FCallOptimization}{\textbf{SendStartOptimizationSignal}}
    \SetKwFunction{FNeedSolve}{\textbf{NeedSolvePGO}}
    \SetKwFunction{FDPGO}{\textbf{DistributedOptimization}}
    \Fn{\FNewkeyframe{${F}_j^t$, $\mathcal{E}_j^t$}} {
        $i \leftarrow$ \FSGM{${F}_j^t$, $\mathcal{E}_j^t$} \\
        $\mathcal{P}_i \leftarrow \mathcal{P}_i \cup{F}_j^t $ \\
        $G_{p_k} \leftarrow G_{p_k} \cup ({F}_j^t, \mathcal{E}_j^t)$ \\
        \ForAll{$r \in \FConnectedRobots{}$}{
            \FFushPartition{$r, {F}_j^t$, $\mathcal{E}_j^t$}
        }
    }
    \Fn{\FRemotekeyframe{${F}_j^t$, $\mathcal{E}_j^t$} } {
        $\mathcal{P}_i \leftarrow \mathcal{P}_i \cup{F}_j^t $ \\
        $G_{p_k} \leftarrow G_{p_k} \cup ({F}_j^t, \mathcal{E}_j^t)$ \\
    }
    \Fn{\FRemotePart{$\mathcal{P}_{new}$} } {
        $\mathcal{P} \leftarrow \mathcal{P}_{new}$ \\
        \FUpdateFENNEL{$\mathcal{P}$} \\
    }
    \Fn{\FLoop} {
        \While {running} {
            \ForAll{$i \in \FConnectedRobots{}$}{
                $\mathcal{F}_{r},\mathcal{E}_{r}  \leftarrow$ \FSyncFullPartition{r}
                $G_{p_k} \leftarrow G_{p_k} \cup (\mathcal{F}_{r},\mathcal{E}_{r} )$ \\
            }

            \If{ \FISMaster{k} and \FNeedRepart{} } {
                $\mathcal{P} \leftarrow $ \FRepart{$G_{p_k}, \mathcal{P}$}

                \ForAll{$r \in \FConnectedRobots{}$}{
                    \FSyncFushPartition{r, $\mathcal{P}$}
                }
                \FUpdateFENNEL{$\mathcal{P}$} \\

            }
            \If {\FNeedSolve{}} {
                \FCallOptimization{}
            }
        }

    }
\end{algorithm}

Alg. \ref{alg:framework} shows the algorithm of our framework, which is individually run on each robot.
The function \textbf{HandleNewKeyframe} is an event-driven function that handles the keyframes and edges extracted by the front-end of the CSLAM system.
Once a new keyframe and its connected edges are detected on each robot, we perform a fast partitioning using FENNEL.
After this, the robot pushes the partitioning result to all robots with the function \textbf{PushKeyframeEdges} with network connections, which are given by the function \textbf{ConnectedRobots}. These robots will handle the keyframe and partitioning result with \textbf{HandleRemoteKeyframe}

As mentioned in Sect. \ref{sect:comm}, a robot swarm using wireless ad hoc network communication may produce changes in the network topology, which we need to take into account when performing repartitioning in the second stage.
\textbf{PartitioningThread} shows the thread that runs the second-stage repartitioning algorithm.
In \textbf{PartitioningThread}, robot $k$ first pull the keyframe, edges that it doesn't have, namely $G_{p_k}-G_{p_r}$, from connected robots $r$ with \textbf{PullDiffPosesAndPartitions}. 
This step is designed to handle the case of multiple sub-swarms merging.
After that, we examine in function \textbf{IsMain} whether the local robot is the main robot of the sub-swarm, where the robot with the smallest ID in each sub-swarm is considered to be the main robot. 
The main robot uses \textbf{NeedRepartition} to determine whether the pose graph maintained by the current sub-swarm needs to be repartitioned. 
Two conditions are taken into account: changes in the network topology and a sufficient number of new keyframes have been received compared to the last repartitioning.  
When either of these conditions is met, the main robot calls \textbf{UnifiedRepartitioning} to perform a repartitioning of the pose graph.
Once the repartitioning is completed, the main robot pushes the new partitions to each of the connected robots by \textbf{PushPartitions} and these robots handle the data with \textbf{HandleRemotePartitions}.
Finally, the main will call \textbf{SendStartOptimizationSignal} to call the swarm or sub-swarm to start solving Eq.\ref{eq:DPGO} and Eq.\ref{eq:DPGO2}, if \textbf{NeedSolvePGO} is satisfied, where a simplified policy is to solve DPGO at a fixed frequency.

In this decentralized design, all keyframes and their attached poses of robots that can communicate are shared among all robots. Still, the robot only uses the poses partitioned to it to create local sub-problems.
Therefore, when some robots disconnect, their partitioned poses are redistributed to the connectable robots.
If a swarm splits into a few sub-swarms, they will perform their own respective  DPGO and will still incorporate all the poses attached to keyframes extracted by the other sub-swarm when it is connectable. 
When the sub-swarms are merged, each robot shares information with other robots, and computational tasks are reallocated to each robot in a balanced manner by BDPGO.

It is notable that the heuristic of FENNEL: 
$ \delta g(v, \mathcal{P}_i) = \left\vert \mathcal{P}_i \cap N(v) \right\vert - \sqrt{k} \frac{n_e}{n_v^{3/2}} \gamma \left\vert \mathcal{P}_i \right\vert^{\gamma-1}$ and the local factor $\nu \frac{n_v}{n_r}$
requires a priori information about the graph with two parameters: $n_e$ and $n_v$, corresponding to the number of vertices (keyframes) and the number of edges of the pose graph.
For SLAM,  $n_e$ and $n_v$ are unavailable since the problem is built incrementally, so we update these two parameters after each repartition.
In function \textbf{UpdateFENNELParam}, $n_e$, $n_v$ is updated according to the size of the current pose graph and the interval frames of the repartition, $dn$, specifically,
        $n_{e_{t+1}} \leftarrow \left\vert V \right\vert + dn$,
        $n_{v_{t+1}} \leftarrow \frac{\left\vert V \right\vert + dn}{\left\vert V \right\vert} \left\vert E \right\vert$.

Another thing to consider is that Alg. \ref{alg:framework} lacks a guarantee that $\mathcal{P}_i$ is internally connected.
However, DGS requires poses' connectivity inside the sub-problem.
To solve this problem, we partition the poses within $\mathcal{P}_i$  into multiple sets based on connectivity, use Dijkstra's algorithm \cite{dijkstra1959note} to search for the shortest path between these sets over the entire scope of $G_p$ and construct new edges using the edges on the shortest path.
This algorithm can fix the connectivity problem and has no significant impact on the system performance.
\begin{figure*}[ht]
    \centering
    \begin{subfigure}[b]{1.0\textwidth}

    \centering
    \settowidth\imagem{\includegraphics[height=3cm]{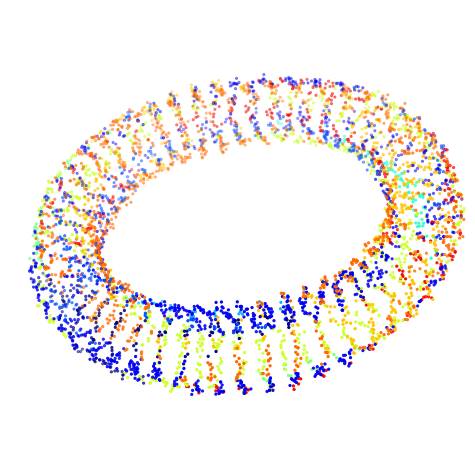}}
    \settowidth\imagem{\includegraphics[height=3cm]{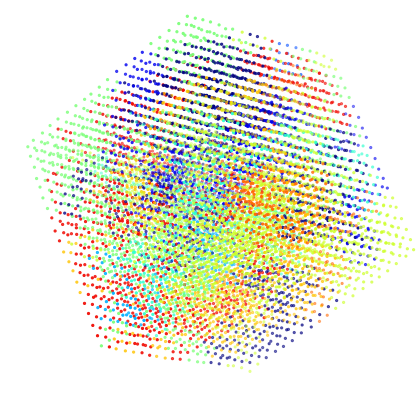}}
    \settowidth\imagem{\includegraphics[height=3cm]{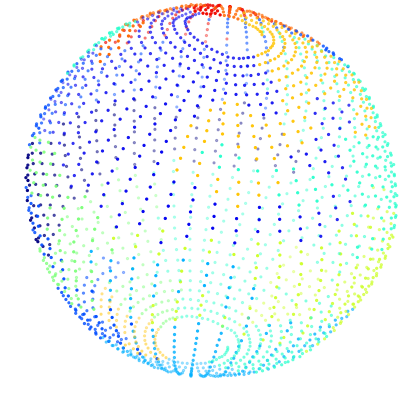}}
    \settowidth\imagem{\includegraphics[height=3cm]{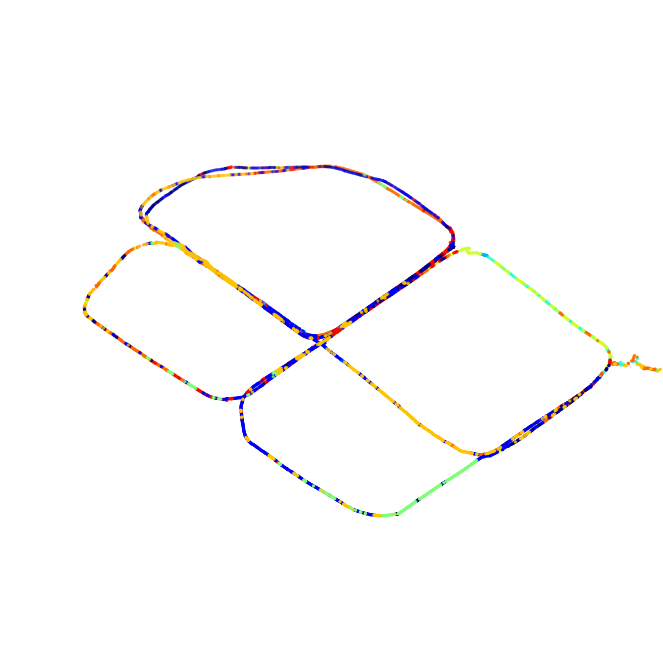}}
    \settowidth\imagem{\includegraphics[height=3cm]{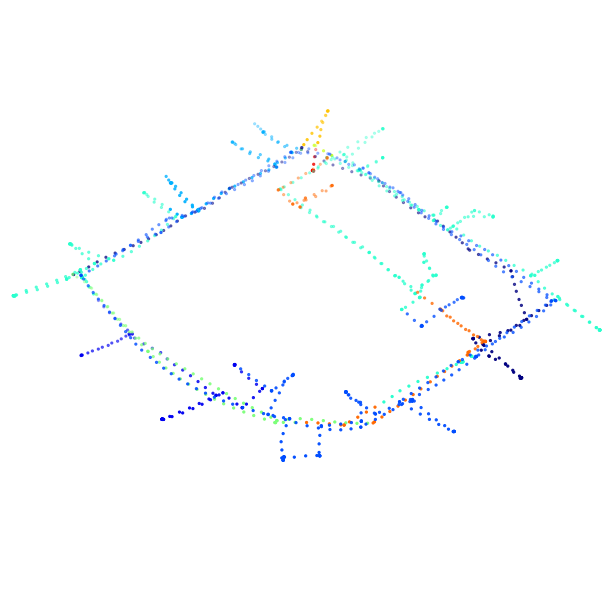}}
    \settowidth\imagem{\includegraphics[height=3cm]{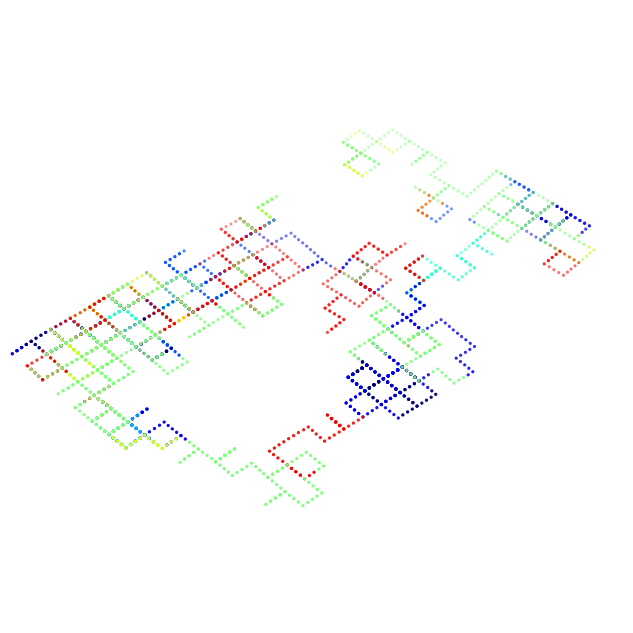}}
    \resizebox{1.0\textwidth}{!}{
        \begin{tabular}{p{\imagem}p{\imagem}p{\imagem}p{\imagem}p{\imagem}p{\imagem}}
    \includegraphics[height=3cm]{generated/torus3D_baseline.png}\newline
    &   \includegraphics[height=3cm]{generated/grid3D_baseline.png}\newline
    &   \includegraphics[height=3cm]{generated/sphere_baseline.png}\newline
    &   \includegraphics[height=3cm]{generated/cubicle_baseline.png}\newline
    &   \includegraphics[height=3cm]{generated/INTEL_baseline.png}\newline
    &   \includegraphics[height=3cm]{generated/M3500_baseline.png}\newline
    \end{tabular}
    }
\end{subfigure}
\begin{subfigure}[b]{1.0\textwidth}
    \centering
    \settowidth\imagem{\includegraphics[height=3cm]{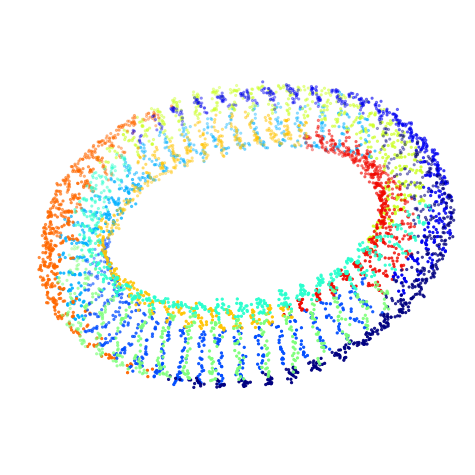}}
    \settowidth\imagem{\includegraphics[height=3cm]{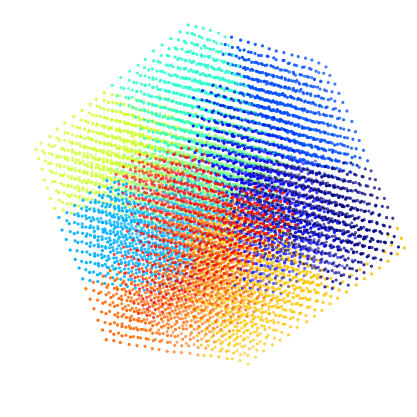}}
    \settowidth\imagem{\includegraphics[height=3cm]{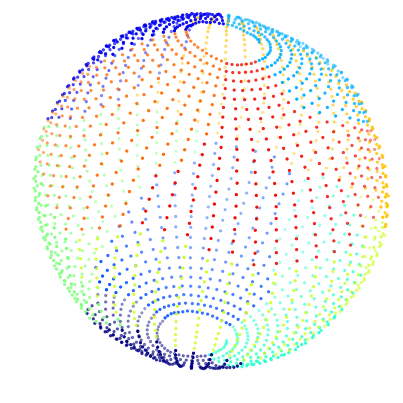}}
    \settowidth\imagem{\includegraphics[height=3cm]{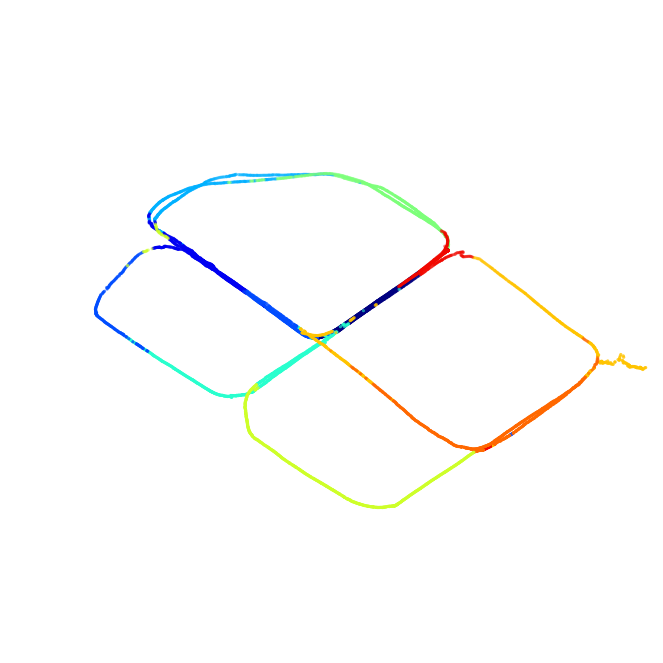}}
    \settowidth\imagem{\includegraphics[height=3cm]{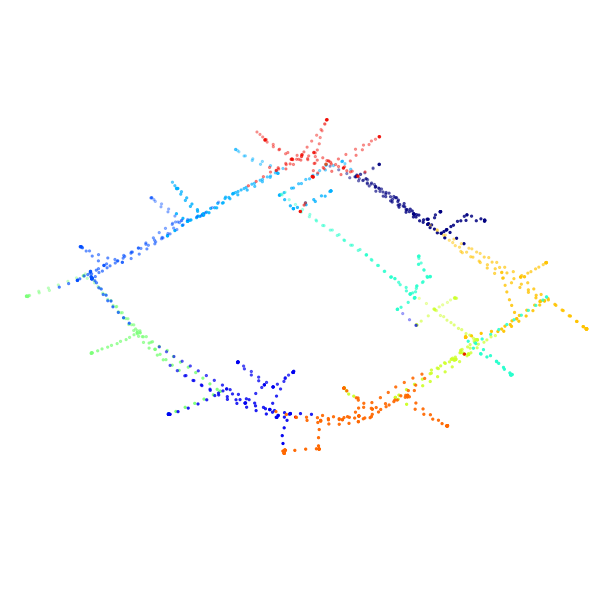}}
    \settowidth\imagem{\includegraphics[height=3cm]{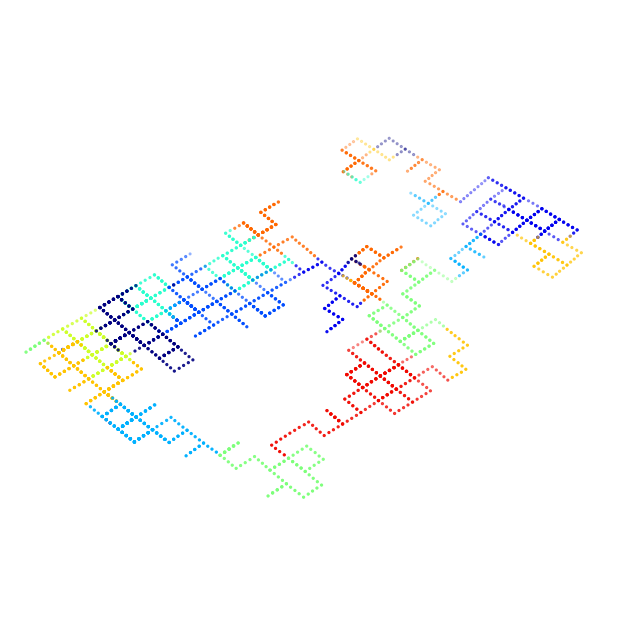}}
    \resizebox{1.0\textwidth}{!}{
        \begin{tabular}{p{\imagem}p{\imagem}p{\imagem}p{\imagem}p{\imagem}p{\imagem}}
    \includegraphics[height=3cm]{generated/torus3D_proposed.png}\newline\subcaption{Torus}
    &   \includegraphics[height=3cm]{generated/grid3D_proposed.png}\newline\subcaption{Grid}
    &   \includegraphics[height=3cm]{generated/sphere_proposed.png}\newline\subcaption{Sphere}
    &   \includegraphics[height=3cm]{generated/cubicle_proposed.png}\newline\subcaption{Cubicle}
    &   \includegraphics[height=3cm]{generated/INTEL_proposed.png}\newline\subcaption{Intel}
    &   \includegraphics[height=3cm]{generated/M3500_proposed.png}\newline\subcaption{Manhattan}
    \end{tabular}
    }
\end{subfigure}
\caption{\small{
    This figure shows the comparison of the graph partitioning results of different methods on the datasets.
    The first row of partitioning is given by the baseline method, and the second row is given by the proposed method.
    Each point in the figures represents a pose labeled with different colors depending on their partitioning. 
The pose graphs are optimized by the proposed method with DGS as the solver. 
       }}\label{fig:plot_partition}
\end{figure*}

\begin{table}[]
    \centering
    \caption{
    \small{\textbf{Comparison of the Partitioning Methods.} $\lambda_{imb}=\frac{\max_{i=1..n}|\mathcal{P}_i|}{\min_{j=1..n}|\mathcal{P}_j|}$ is the imbalance factor, while $\bar\lambda_{imb}$ is the average imbalanced factor. 
    $\lambda_{cut}=\frac{|E_{cut}|}{|V|}$ is the edge-cut factor, and $\bar\lambda_{cut}$ is its average value.
    $\lambda_{vol}=\frac{comm(\mathcal{P})}{|V|}$ is the communication volume factor and its average value is $\bar\lambda_{vol}$.
    The overhead is the average time cost for graph partitioning per keyframe.
    }
     }
    \label{tab:parititions}
    \setlength\tabcolsep{3pt}
    \begin{tabular}{c|c|c|c|c|c|c|c}
    \hline
    \multirow{2}{*}{Dataset}   & \multirow{2}{*}{\begin{tabular}[c]{@{}c@{}}Poses/\\ Edges\end{tabular}} & \multirow{2}{*}{Partition} & \multirow{2}{*}{$\bar{\lambda}_{imb}$} & \multirow{2}{*}{$\bar{\lambda}_{cut}$} & \multirow{2}{*}{$\bar{\lambda}_{vol}$} & \multicolumn{2}{c}{Overhead} \\ \cline{7-8} 
                               &                                                                         &                            &                      &                      &                      & PC                      &TX2 \\ \hline\hline
    \multirow{3}{*}{Torus}     & \multirow{3}{*}{\begin{tabular}[c]{@{}c@{}}5000/\\ 9162\end{tabular}}   & Baseline                   & 26.91                & 0.17                 & 0.30                 & -                       & -    \\ \cline{3-8} 
                               &                                                                         & NoFENNEL                   & 1.42                 & 0.09                 & 0.16                 & 1.9ms                   &10.5ms\\ \cline{3-8} 
                               &                                                                         & \textbf{Proposed}          & \textbf{1.21}        & \textbf{0.08}        & \textbf{0.15}        & 1.9ms                   &10.4ms\\ \hline\hline
    \multirow{3}{*}{Grid}      & \multirow{3}{*}{\begin{tabular}[c]{@{}c@{}}8000/\\ 22294\end{tabular}}  & Baseline                   & 18.82                & 0.54                 & 0.76                 & -                       &  -   \\ \cline{3-8} 
                               &                                                                         & NoFENNEL                   & 1.31                 & 0.15                 & 0.25                 & 4.1ms                   &21.7ms\\ \cline{3-8} 
                               &                                                                         & \textbf{Proposed}          & \textbf{1.21}        & \textbf{0.13}        & \textbf{0.23}        & 4.2ms                   &21.6ms\\ \hline\hline
    \multirow{3}{*}{Sphere}    & \multirow{3}{*}{\begin{tabular}[c]{@{}c@{}}2200/\\ 8662\end{tabular}}   & Baseline                   & 6.54                 & 0.38                 & 0.38                 & -                       & -   \\ \cline{3-8} 
                               &                                                                         & NoFENNEL                   & 1.43                 & 0.29                 & 0.31                 & 1.7ms                   &8.7ms\\ \cline{3-8} 
                               &                                                                         & \textbf{Proposed}          & \textbf{1.30}        & \textbf{0.24}        & \textbf{0.26}        & 1.7ms                   &8.7ms\\ \hline\hline
    \multirow{3}{*}{Cubicle}   & \multirow{3}{*}{\begin{tabular}[c]{@{}c@{}}5750/\\ 12593\end{tabular}}  & Baseline                   & 64.74                & 0.31                 & 0.46                 & -                       &-    \\ \cline{3-8} 
                               &                                                                         & NoFENNEL                   & 1.39                 & 0.10                 & 0.17                 & 2.4ms                   &13.5ms\\ \cline{3-8} 
                               &                                                                         & \textbf{Proposed}          & \textbf{1.22}        & \textbf{0.08}        & \textbf{0.14}        & 2.5ms                   &13.6ms\\ \hline\hline

    \multirow{3}{*}{Intel}     & \multirow{3}{*}{\begin{tabular}[c]{@{}c@{}}1228/\\ 1501\end{tabular}}   & Baseline                   & 57.81                & 0.11                 & \textbf{0.16}        & -                       & -   \\ \cline{3-8} 
                               &                                                                         & NoFENNEL                   & 2.63                 & 0.10                 & 0.18                 & 0.4ms                   &2.3ms\\ \cline{3-8} 
                               &                                                                         & \textbf{Proposed}          & \textbf{1.39}        & \textbf{0.10}        & 0.19                 & 0.4ms                   &2.4ms\\ \hline\hline
    \multirow{3}{*}{Manhattan} & \multirow{3}{*}{\begin{tabular}[c]{@{}c@{}}3500/\\ 5548\end{tabular}}   & Baseline                   & 82.96                & 0.17                 & 0.29                 & -                       & -   \\ \cline{3-8} 
                               &                                                                         & NoFENNEL                   & 1.64                 & 0.08                 & 0.15                 & 1.2ms                   &6.8ms\\ \cline{3-8} 
                               &                                                                         & \textbf{Proposed}          & \textbf{1.23}        & \textbf{0.07}        & \textbf{0.14}        & 1.3ms                   &7.0ms\\ \hline\hline
    \end{tabular}
    \end{table}
\section{Experiments}\label{sect:evaluation}
A robot swarm in the real-world environment gradually builds pose graphs as they move.
To simulate this motion process reasonably well on datasets, we use the Multiple Traveling Salesman Problem Solver inside LKH-3\cite{helsgaun2017extension} to solve 10 random paths on the original pose graph datasets as the robot paths. 
These 10 paths are subsequently assigned to timestamps uniformly as a simulation of the motion of the 10 robots.
The edges along the paths can be treated as a simulation of the odometry measurements.
In the simulation of the proposed framework, these keyframes are sequentially sent to BDPGO for each robot according to the timestamps, and the pose graph is built incrementally.
We also compare our method with a method that removes the first stage, hereafter called the NoFENNEL method, to demonstrate the necessity of introducing the first-stage FENNEL method.
In the framework, the Unified Repartitioning Algorithm is implemented using the ParMETIS\footnote{http://glaros.dtc.umn.edu/gkhome/metis/parmetis/overview} library.

\subsection{Evaluation of Two-stage Graph Partitioning}\label{sect:expr_part}

\begin{table*}[]
    \centering
    \caption{
\small{
    \textbf{Comparison of Proposed Framework and Baseline Method with DGS Optimization.}
Avg. Add. Edges are the average number of additional edges generated to solve the internal connectivity issue of the partitions.
\textit{Avg. Time} is the time of DGS solving the problem, which is determined by the slowest robot.
\textit{Avg. Util. Rate} is the average utilization rate of robots in solving the DPGO, which is defined as $\sum_i t_{s_i}/(n\cdot t_{s_m})$, where $t_{s_i}$ is the solving time of robot $i$, and $t_{s_m}$ is the solving time of the slowest robot. 
\textit{Total Comm.} is the sum of the communication volume of each iteration during the whole process.
}}
    \label{tab:dgs_benchmark}
    \begin{tabular}{c|c|c|c|c|c|c|c|c}
    \hline
    Dataset                    & Partitioning      &Avg. Add. Edges & Avg. Time        & Avg. Util. Rate& Avg. Iters.  & Avg. Initial Cost & Avg. Final Cost & Total Comm.      \\ \hline\hline
    \multirow{4}{*}{Torus}     & Baseline          &-                & 272.5ms          &    47.0\%      & 56.9         & 11462.12          & 37.59           & 6466041          \\ \cline{2-9} 
                               & NoFENNEL          &8.7              & 113.8ms          &    86.7\%      & 44.6         & 11947.34          & 37.99           & 1932129          \\ \cline{2-9} 
                               & \textbf{Proposed} &\textbf{2.7}     & \textbf{90.6ms}  &\textbf{88.1\%} & \textbf{35.6}& 11618.69          & \textbf{37.54}  & \textbf{1440603} \\ \hline\hline

    \multirow{4}{*}{Grid}      & Baseline          &-                & 614.0ms          &    47.4\%      & 67.0         & 318525.29         & 118.94          & 46983149         \\ \cline{2-9}  
                               & NoFENNEL          &10.1             & 443.7ms          &    75.3\%      & 54.6         & 323180.48         & 119.81          & 11840985       \\ \cline{2-9} 
                               & \textbf{Proposed} &\textbf{2.1}     & \textbf{299.3ms} &\textbf{75.0\%} & \textbf{42.3}& 319925.99         & \textbf{118.31} & \textbf{6469528}\\ \hline\hline
    
    \multirow{4}{*}{Sphere}    & Baseline          &-                & 352.5ms          &    46.8\%      & 98.4         & 36803739.59       & 58562.76        & 2788266          \\ \cline{2-9}  
                               & NoFENNEL          &7.6              &\textbf{186.2ms}  &    84.0\%      &\textbf{100.2}& 37199866.58       & 112464.94       & 1942471  \\ \cline{2-9} 
                               & \textbf{Proposed} &\textbf{2.8}     & 186.7ms          &\textbf{85.1\%} & 103.5        & 36945114.72       &\textbf{44052.36}& \textbf{1529121}  \\ \hline\hline
    
    \multirow{4}{*}{Cubicle}   & Baseline          &-                & 133.5ms          & 45.3\%         & 32.6         & 33846.0           & 2.00            & 4934576           \\ \cline{2-9} 
                               & NoFENNEL          &9.9              & 66.4ms           & \textbf{79.9\%}& 16.0         & 34888.0           & 1.61            & 1081773           \\ \cline{2-9} 
                               & \textbf{Proposed} &\textbf{2.6}     & \textbf{62.7ms}  & 79.6\%         & \textbf{14.8}& 33980.7           &\textbf{1.57}    & \textbf{799251}  \\ \hline\hline

    \multirow{4}{*}{Intel}     & Baseline          &-                & 74.9ms           &  32.7\%        & 38.5         & 10308.29          &0.25             & 165259           \\ \cline{2-9} 
                               & NoFENNEL          &8.6              & 19.1ms           &  71.7\%        & 25.6         & 11111.39          &0.22             & 94970            \\ \cline{2-9} 
                               & \textbf{Proposed} &\textbf{6.1}     & \textbf{13.4ms}  &\textbf{77.3}\% & \textbf{18.7}& 10887.91          &\textbf{0.21}    & \textbf{62981}   \\ \hline\hline

    \multirow{4}{*}{Manhattan} & Baseline          &-                & 216.6ms          & 38.0\%         & 66.7         & 36638.00          & 59.59           & 4023810          \\ \cline{2-9} 
                               & NoFENNEL          &8.5              & 100.3ms          & 85.4\%         & 67.0         & 39088.62          & 1.24            & 1557413  \\ \cline{2-9} 
                               & \textbf{Proposed} &\textbf{3.4}     & \textbf{92.2ms}  &\textbf{86.7\%} & \textbf{63.8}& 37794.02          & \textbf{1.21}   & \textbf{1279589} \\ \hline\hline
                               
\end{tabular}
\end{table*}

\begin{table}[]
    \centering
    \caption{
            \small{
                \textbf{Comparison of Proposed Framework and Baseline Method with ASAPP Optimization.}
                The ratio of the minimum number of iterations to the maximum number of iterations is recorded as \textit{Iter. Imb. Rate}, which shows the imbalance level of the iterations on different robots.
                Comm. is the total communication volume during the whole process.
            }}
    \label{tab:asapp_benchmark}
    \setlength\tabcolsep{5pt}
    \begin{tabular}{c|c|c|c|c|c}
    \hline
    \multirow{2}{*}{Dataset}   & \multirow{2}{*}{Partitioning} & \multirow{2}{*}{\begin{tabular}[c]{@{}c@{}}Iter. Imb. \\ Rate\end{tabular}} & \multicolumn{2}{c|}{Cost}                                 & \multirow{2}{*}{Comm.} \\ \cline{4-5}
                               &                               &                                                                             & \multicolumn{1}{l|}{Initial} & \multicolumn{1}{l|}{Final} &                             \\ \hline\hline
    \multirow{3}{*}{Torus}     & Baseline                      & 13.5\%                                                                      & \multirow{3}{*}{11.06}       & 10.10                      & 60070885                    \\ \cline{2-3} \cline{5-6} 
                               & NoFENNEL                      & 82.1\%                                                                      &                              & 9.76                       & 29527850                    \\ \cline{2-3} \cline{5-6} 
                               & \textbf{Proposed}             & \textbf{82.7\%}                                                             &                              & \textbf{9.76}              & \textbf{27052625}           \\ \hline\hline
    \multirow{3}{*}{Grid}      & Baseline                      & 11.8\%                                                                      & \multirow{3}{*}{25.31}       & 23.72                      & 187145260                   \\ \cline{2-3} \cline{5-6} 
                               & NoFENNEL                      & 80.0\%                                                                      &                              & 23.57                      & 63269096                    \\ \cline{2-3} \cline{5-6} 
                               & \textbf{Proposed}             & \textbf{80.4\%}                                                             &                              & \textbf{23.57}             & \textbf{56649352}           \\ \hline\hline
    \multirow{3}{*}{Sphere}    & Baseline                      & 26.1\%                                                                      & \multirow{3}{*}{413.97}      & 98.28                      & 19546675                    \\ \cline{2-3} \cline{5-6} 
                               & NoFENNEL                      & 79.9\%                                                                      &                              & 97.19                      & 15276149                    \\ \cline{2-3} \cline{5-6} 
                               & \textbf{Proposed}             & \textbf{80.8\%}                                                             &                              & \textbf{97.15}             & \textbf{12421322}           \\ \hline\hline
    \multirow{3}{*}{Cubicle}   & Baseline                      & 11.6\%                                                                      & \multirow{3}{*}{2.80}        & 1.17                       & 88970937                    \\ \cline{2-3} \cline{5-6} 
                               & NoFENNEL                      & \textbf{80.1\%}                                                             &                              & 0.99                       & 35216805                    \\ \cline{2-3} \cline{5-6} 
                               & \textbf{Proposed}             & 79.7\%                                                                      &                              & \textbf{0.99}              & \textbf{27999097}           \\ \hline\hline
    \multirow{3}{*}{Intel}     & Baseline                      & 31.1\%                                                                      & \multirow{3}{*}{0.40}        & 0.28                       & \textbf{7378052}            \\ \cline{2-3} \cline{5-6} 
                               & NoFENNEL                      & 75.7\%                                                                      &                              & 0.28                       & 8205536                     \\ \cline{2-3} \cline{5-6} 
                               & \textbf{Proposed}             & \textbf{80.7\%}                                                             &                              & \textbf{0.28}              & 8046855                     \\ \hline\hline
    \multirow{3}{*}{Manhattan} & Baseline                      & 14.8\%                                                                      & \multirow{3}{*}{48.86}       & 3.95                       & 50002552                    \\ \cline{2-3} \cline{5-6} 
                               & NoFENNEL                      & 81.8\%                                                                      &                              & 3.79                       & 23948958                    \\ \cline{2-3} \cline{5-6} 
                               & \textbf{Proposed}             & \textbf{83.3\%}                                                             &                              & \textbf{3.79}              & \textbf{21740259}           \\ \hline
    \end{tabular}
    \end{table}
A visualization of the pose graph partitioned by the baseline method and proposed method can be found in Fig. \ref{fig:plot_partition}.
In the evaluations, the repartitioning duration $dn$ is set to 100, $\gamma$  and $\nu$ in FENEEL's heuristics function is set to 1.5 and 1.1.
An intuitive impression is that the partitioning results of the baseline method are messy in Fig. \ref{fig:plot_partition}  because they are randomly generated paths on the graph, while the partitions generated by the proposed method are well organized.
Well-organized partitioning implies a more balanced partition with less communication.
This intuitive impression can be confirmed in Table \ref{tab:parititions}.
The results clearly show that the proposed method has better balance and produces results with a lower communication volume than the baseline and NoFENNEL methods throughout the incremental creation of the pose graph.

Finally, to show the more realistic performance, we measure the average overhead of the partitioning method on both a PC (with AMD R9 5900X CPU) and an embedded platform, NVIDIA Jetson TX2.
Even on the embedded device, the overhead brought by the proposed method is acceptable.
The overhead is similar between the proposed method and NoFENNEL since FENNEL only requires computing a heuristic. In contrast, the proposed method has better performance on balancing and minimizing communication volume.
On the other hand, as a comparison, it takes up to 203.0ms (on Grid3D dataset) if METIS is used to partition the pose graph on the PC and up to 947.2ms on the TX2 while achieving similar results compared to the proposed method, which is unacceptable for a real-time system.
This comparison validates the necessity of our proposed two-stage approach for the incremental CSLAM problem other than using direct graph partitioning.

\subsection{Evaluation of BDPGO with Solvers}\label{sect:expr_dpgo}

Beyond the performance of the graph partitioning itself, the final question we need to face is how exactly these partitionings affect the final performance of PGO.
In this paper, we test two representative DPGO methods, synchronous DGS and asynchronous ASAPP, and present impressive results, as shown in Table \ref{tab:dgs_benchmark} and Table \ref{tab:asapp_benchmark}. 
We modify distributed-mapper\footnote{https://github.com/CogRob/distributed-mapper} for testing the DGS solver and implement ASAPP in C++ with the help of the Ceres Solver\cite{ceres-solver} for automatic differentiation and the sparse data structures.
In the benchmark, we use the three partitioning methods described in Sect. \ref{sect:expr_part} to partition the pose graph to generate local subproblems according to Eq.\ref{eq:DPGO} and Eq. \ref{eq:DPGO2}.
During the process, we solve DPGO for every ten new keyframes that are added, and the average performance is shown in Table \ref{tab:dgs_benchmark} and Table \ref{tab:asapp_benchmark}.
The benchmarks are performed on a powerful PC with an AMD R9 5900X CPU.
To show the convergence performance of the algorithm, we use the raw values in the dataset to initialize the pose graph every time we solve DPGO.
The maximum iteration number of DGS is limited to 100 for both rotation initialization and pose optimization, and the optimization result is shown in Fig. \ref{fig:plot_partition}.

We can find that the DGS incorporating the proposed method is 2 to 5 times faster than the baseline method in the majority of cases in Table \ref{tab:dgs_benchmark} and requires minimum iterations.
However, in the Sphere dataset, the NoFENNEL method is slightly faster than the proposed method. Nevertheless, we can find that the final cost of the NoFENNEL method is much larger than that of the proposed method in this dataset, which means that the NoFENNEL method leads to suboptimality.
Beyond the Sphere dataset, we can also find that the proposed method gives the smallest final cost in all other datasets.
In some datasets, the proposed method delivers results that are even orders of magnitude smaller, which means that the proposed partitioning brings the best convergence property.
What's more, the key idea in this paper is to speed up the DPGO by increasing the utilization of the computation of each robot.
A bigger utilization rate means the robot takes less time to wait for exchanging poses but more time to solve the subproblem.
This is also verified in Table \ref{tab:dgs_benchmark}, where the utilization rate of the proposed method is the highest on most of the datasets, which leads to a fast optimization speed.

Furthermore, additional edges in Table \ref{tab:dgs_benchmark} are generated from existing edges to fix internal connectivity, which may lead to the problem of double-counting of information and accuracy degradation, so fewer additional edges point to a better partitioning method.
On all datasets, the proposed method generates fewer additional edges than the NoFENNEL method. 
This is because the
$\left\vert \mathcal{P}_i \cap N(v) \right\vert$ item in FENNEL's heuristic leads to better connectivity of the partitioning.
Notably, the additional edges slightly increase the initial costs of the proposed method, while the final cost is smaller. 
This verifies that these additional edges have no significant effect on the proposed method.

In addition, to verify the performance of the BDPGO framework using an asynchronous DPGO, we performed a verification using ASAPP, which is set to run for 5 seconds in each time of optimization. 
The gradient descent rate is set to $10^{-5}$ over all datasets, and we initialized the pose graph using DGS as suggested in Tian et al. \cite{tian2020asynchronous}.
Table \ref{tab:asapp_benchmark} shows the comparison of different partitioning approaches on ASAPP.
We can find that the final costs of the proposed method and the NoFENNEL method are close, which indicates that they have similar convergence performance, and both have better convergence performance than the baseline method.
What's more, we find that the imbalance rate of the baseline method is bigger than the proposed method, which also results in wasting computational resources.
\begin{figure}[t]
    \begin{subfigure}[b]{1.0\linewidth}
    
        \centering
        \settowidth\images{\includegraphics[height=3cm]{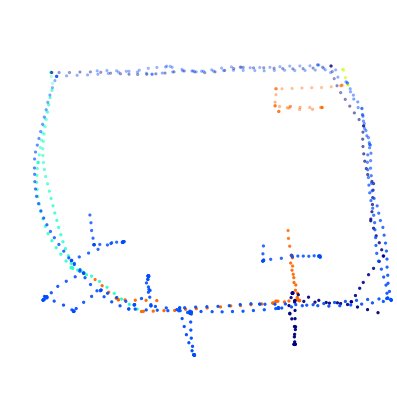}}
        \settowidth\images{\includegraphics[height=3cm]{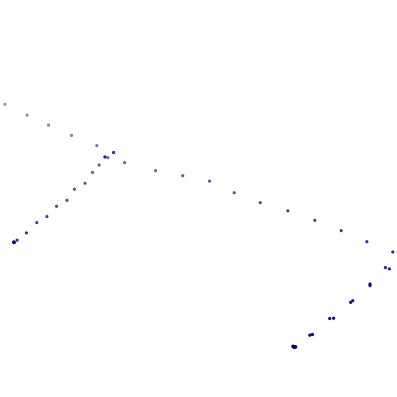}}
        \settowidth\images{\includegraphics[height=3cm]{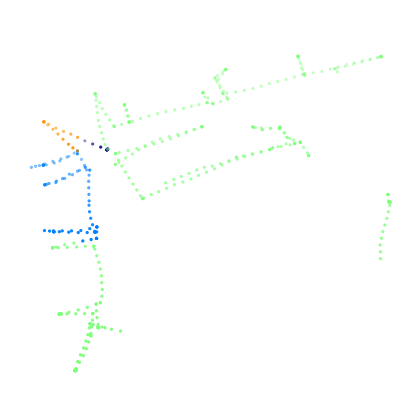}}
        \resizebox{1.0\linewidth}{!}{
            \begin{tabular}{p{\images}p{\images}p{\images}}
        \includegraphics[height=3cm]{baseline_time374_subswarm_0.png}\newline
        &   \includegraphics[height=3cm]{baseline_time374_subswarm_1.png}\newline
        &   \includegraphics[height=3cm]{baseline_time374_subswarm_3.png}\newline
        \end{tabular}
        }
        \subcaption{\small{Optimized pose graph with sub-swarms of baseline method}}
    \end{subfigure}
    \begin{subfigure}[b]{1.0\linewidth}
    
        \centering
        \settowidth\images{\includegraphics[height=3cm]{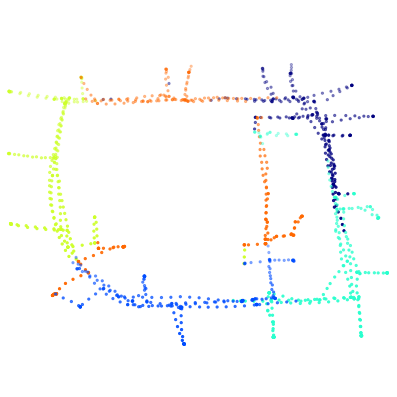}}
        \settowidth\images{\includegraphics[height=3cm]{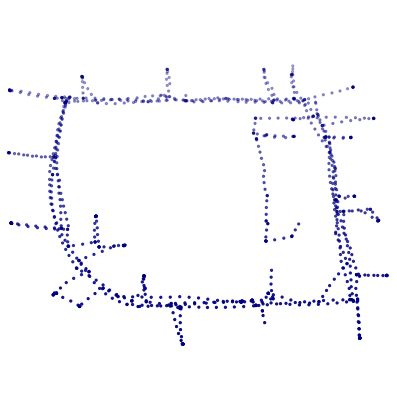}}
        \settowidth\images{\includegraphics[height=3cm]{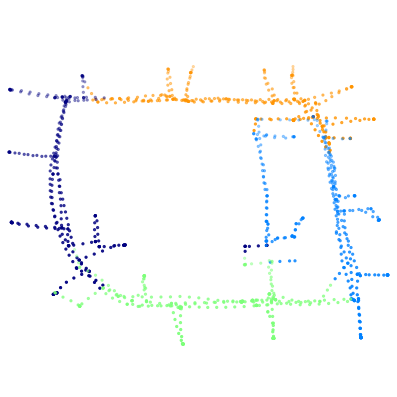}}
        \resizebox{1.0\linewidth}{!}{
            \begin{tabular}{p{\images}p{\images}p{\images}}
        \includegraphics[height=3cm]{proposed_time374_subswarm_0.png}\newline
        &   \includegraphics[height=3cm]{proposed_time374_subswarm_1.png}\newline
        &   \includegraphics[height=3cm]{proposed_time374_subswarm_3.png}\newline
        \end{tabular}
        }
        \subcaption{\small{Optimized pose graph with sub-swarms of proposed method}}
    \end{subfigure}
    \caption{\small{
        This figure compares the proposed method and the baseline method when the swarm is split into multiple sub-swarms due to communication problems. 
        On each sub-swarm, the poses belonging to different partitions are labeled with different colors.
    }}\label{fig:posegraph_adhoc}
    \end{figure}

One of the major advantages of BDPGO is the smaller volume of communication required during optimization, which is also verified in Table \ref{tab:dgs_benchmark} and Table \ref{tab:asapp_benchmark}.
On most datasets, the proposed method has a lower communication volume than the NoFENNEL method and much lower than the baseline method.
Some readers may think that the proposed method leads to communication volume wasting during graph partitioning. Namely, the baseline method does not require each robot to have all the keyframes of the entire swarm, but only the remote poses connected to its poses when performing DPGO. In contrast, the proposed method needs robots to pull all keyframes in the entire swarm.
However, for the vast majority of CSLAM systems\cite{lajoie2020door,zou2019collaborative}, robots use loop closure detection other than object-based approaches to detect relationships between keyframes, which requires robots to send all their keyframes to the swarm to maximize the re-call rate.
    
Another communication-related advantage of BDPGO is the robustness to partial robots in the swarm loss of connection, which can be caused by robot failure or network topology changes.
Fig. \ref{fig:posegraph_adhoc} shows an example of simulation of robot swarm with wireless ad hoc on the Intel dataset. 
In the simulation, the communication distance of the robot is limited to 10 m, which causes the swarm to be split into three subswarms that cannot communicate with each other at the end of the task.
As shown in the figure, with the baseline method, each of the three sub-swarms has a pose graph that is part of the global pose graph.
Nevertheless, the network topology does not always remain the same during robot movement, which allows the BDPGO to exchange most of the poses from other robots throughout the process of sub-swarm splitting and merging.
The pose graph for each sub-swarm using BDPGO is more complete compared to the baseline, and the coherence of the poses is kept.
This coherence also holds for partial robot failures, in which case the equivalent of the sub-swarm in Fig. \ref{fig:posegraph_adhoc} will not reconnect again.
In addition, we found in the simulation that partial robot disconnections may cause the pose graph of the sub-swarm with the baseline method to lose connectivity. 
For example, in the pose graph, the poses of robot 1 is connected to poses of robot 2, and poses robot 2 is connected to poses of robot 3. when robot 2 is lost, we can't solve the DPGO using the baseline method since the pose graph is disconnected.
Conversely, the proposed method eliminates this problem because the poses and edges partitioned to robot 2 in the above example are redistributed to robot 1 and robot 3.

\section{Conclusion and Future Works}
In this paper, we propose BDPGO, a balanced DPGO framework.
With this approach, we significantly speed up existing DPGO methods while reducing the communication volume and being capable of coping with wireless network changes, robot failures.
This framework can effectively improve CSLAM with DPGO for robot swarms working in real-world environments. 
In addition, although only DPGO is discussed in this paper, our proposed approach can actually be extended to all CSLAM problems that require distributedly solving the factor graph.
We believe that the ideas embedded in this paper, namely, decoupling the sensor and computational functions on robot clusters and maximizing the utilization of resources in robot swarms through comprehensive scheduling, will inspire more research on robot swarms.

\bibliographystyle{IEEEtran}
\bibliography{hao2021} 

\end{document}